\documentclass[twocolumn,times,final,numbers]{elsarticle}

\usepackage{prletters}
\usepackage{framed,multirow}
\usepackage{comment}

\usepackage{amssymb}
\usepackage{latexsym}

\usepackage{url}
\usepackage{xcolor}

\usepackage{ulem}

\definecolor{newcolor}{rgb}{.8,.349,.1}

\begin{document}

\begin{comment}
    
\begin{table*}[!th]

\section*{Graphical Abstract (Optional)}
To create your abstract, please type over the instructions in the
template box below.  Fonts or abstract dimensions should not be changed
or altered. 

\vskip1pc
\fbox{
\begin{tabular}{p{.4\textwidth}p{.5\textwidth}}
\bf Attention-Based Segmentation of WMHs and Differentiation of Vascular vs. Demyelinating Lesions  \\
Aina Tur-Serrano, Gabriel Moyà-Alcover, Francisco J. Perales López \\[1pc]
\includegraphics[width=0.95\textwidth]{GraphicalAbstract.pdf}
& 

%}\\
\end{tabular}
}

\end{table*}

\clearpage
\thispagestyle{empty}

\vspace*{-1pc}

\begin{table*}[!t]

\section*{Research Highlights (Required)}

\vskip1pc

\fboxsep=6pt
\fbox{
\begin{minipage}{.95\textwidth}
It should be short collection of bullet points that convey the core
findings of the article. It should  include 3 to 5 bullet points
(maximum 85 characters, including spaces, per bullet point.)  
\vskip1pc
\begin{itemize}

 \item \textcolor{black}{WMH segmentation using attention modules.}

 \item \textcolor{black}{Morphological feature extraction for lesion differentiation.}

 \item \textcolor{black}{Classification using ground truth and segmentation outputs.} 

 \item \textcolor{black}{Integration of five public datasets for generalizability.}

\end{itemize}
\vskip1pc
\end{minipage}
}

\end{table*}

\clearpage
\end{comment}

\setcounter{page}{1}

\begin{frontmatter}

\title{Attention-Based Segmentation of WMHs and Differentiation of Vascular vs. Demyelinating Lesions}

\author[1]{Aina \surname{Tur-Serrano}\corref{cor1}} 
%\cortext[cor1]{Corresponding author: 
 % Tel.: +34-971-172-711;}
%\ead{aina.tur@uib.cat}

\author[1,2]{Gabriel \surname{Moyà-Alcover}}
\author[1]{Francisco J. \surname{Perales López}}

%% \affiliation[label1]{organization={},%Department and Organization
%%             addressline={},
%%             city={},
%%             citysep={}, % use if no comma needed between city and postcode%%             
%%             postcode={},
%%             state={},
%%             country={}}

\affiliation[1]{organization={UGiVIA Research Group, Dpt. of Mathematics and Computer Science, Universitat de les Illes Balears},
                city={Palma de Mallorca}, 
                postcode={07122}, 
                country={Spain}}

\affiliation[2]{organization={Laboratory for Artificial Intelligence Applications (LAIA@UIB), Dpt. of Mathematics and Computer Science, Universitat de les Illes Balears},
                city={Palma de Mallorca}, 
                postcode={07122}, 
                country={Spain}}

\received{1 May 2013}
\finalform{10 May 2013}
\accepted{13 May 2013}
\availableonline{15 May 2013}
\communicated{S. Sarkar}

\begin{abstract}
White Matter Hyperintensities (WMHs) are commonly observed in brain Magnetic Resonance \textcolor{black}{Imaging} (MRI) scans. \textcolor{black}{They} are associated with various neurological conditions, including vascular and inflammatory demyelinating diseases. Despite differing in etiology, WMHs from these conditions often appear similar on \textcolor{black}{Fluid Attenuated Inversion Recovery (FLAIR)} images. \textcolor{black}{This similarity makes} differential diagnosis challenging. \textcolor{black}{In this work, we highlight} %%work highlights%%
the potential of combining attention-based segmentation with feature-driven classification. \textcolor{black}{This approach supports} %%to support%%
more accurate and efficient \textcolor{black}{classification} %differentiation%
between vascular and demyelinating white matter pathologies.
For segmentation, we evaluate the effectiveness of attention mechanisms, \textcolor{black}{specifically the Bottleneck Attention Module (BAM) and the Convolutional Block Attention Module (CBAM). We also test different architectures, particularly Attention U-Net. In addition, we explore advanced training strategies, such as patch-based learning and a 2.5D approach, to enhance lesion detection. }

\textcolor{black}{After} segmentation, we extract morphological features from the lesion masks. \textcolor{black}{We then} use them to classify WMHs based on their underlying cause. Our experiments utilize five publicly available datasets with diverse imaging protocols to promote model generalizability, despite limited sample sizes. The results suggest that attention-based segmentation and feature-driven classification offer a promising direction for \textcolor{black}{discriminating} % the differentiation of% 
vascular and demyelinating white matter lesions. \textcolor{black}{Further validation in larger clinical cohorts is still needed. }%%though validation is needed in larger clinical cohorts.
\end{abstract}


\end{frontmatter}

\section{Introduction}
\label{sec1}
White Matter Hyperintensities (WMHs) are abnormal areas of signal intensity on brain Magnetic Resonance Imaging (MRI). They appear bright on T2-weighted (T2), \textcolor{black}{Proton Density-Weighted} (PD) and Fluid-Attenuated Inversion Recovery (FLAIR) sequences. \textcolor{black}{Among these, FLAIR is}
the most sensitive imaging modality for detecting these changes in the white matter \cite{wardlaw2015wmh}. 

In this study, we focus on two common causes of white matter diseases: vascular causes and inflammatory demyelinating conditions. Although these diseases \textcolor{black}{have distinct} origins, their white matter lesions often \textcolor{black}{share} common imaging features. \textcolor{black}{These include} similar shapes and locations, especially periventricular \textcolor{black}{lesions}. This overlap can complicate differential diagnosis and \textcolor{black}{pose} a challenge in routine clinical settings \citep{zhang2023white}.

Early detection of WMHs is crucial, as it helps reduce the risk of disease progression. \textcolor{black}{Moreover, it} enables timely preventive interventions that may slow or even reverse early brain damage \citep{wardlaw2015wmh}. However, manual annotations of these lesions can be time-consuming and subjective, highlighting the need for reliable automated tools.

%To address this, we utilize deep learning models with attention mechanisms, \textcolor{black}{including} the Attention U-Net architecture \textcolor{black}{as well as other attention modules}. 
\textcolor{black}{To tackle this problem, we leverage deep learning models with attention mechanisms, including Attention U-Net \citep{attentionunet} and other attention-based modules such as the Bottleneck Attention Module (BAM) \citep{BAM} and the Convolutional Block Attention Module (CBAM) \citep{CBAM}.}
%These modules help improve lesion localization, particularly for small or subtle lesions. 
\textcolor{black}{We expect these modules to enhance both spatial and channel feature representation and improve lesion delineation. Recent studies have demonstrated their effectiveness in various medical image segmentation tasks \citep{bamapp, cbamapp}, yet the use of these modules for WMH segmentation remains unexplored}. We adopted different approaches such as 2D slices,  patch-based training and 2.5D input. \textcolor{black}{In particular, the latter uses} adjacent slices as input to provide limited spatial context while preserving computational efficiency. 

Beyond segmentation, a key objective of this work is to explore classification of WMHs based on their underlying origin. 
We evaluated feature-based classification, where morphological descriptors are extracted from segmented lesion masks. \textcolor{black}{These descriptors are then} used to train a classifier. This two-step approach enhances clinical interpretability\textcolor{black}{:} segmentation provides spatial and morphological information about the lesions, while \textcolor{black}{the extracted features enable} a more targeted analysis of lesion shape, size and distribution. \textcolor{black}{The work of Ali et al. \cite{corneal} strongly motivates this direction, showing the value of combining segmentation with interpretable classification in diagnostic workflows.}

Our work exclusively analyzes WMHs visible on FLAIR sequences. We include five publicly available datasets\textcolor{black}{,} with varying imaging protocols and scanner types, aiming to develop more generalizable models.

\section{Methodology} \label{sec:methods}
\textcolor{black}{Following best practices for clear and concise methodological reporting in medical image analysis, as demonstrated by Al-Quraishi et al. \citep{datasetsdesc}, who provided transparent and structured descriptions of datasets, preprocessing steps, feature selection strategies, and classifier design for tumor viability prediction, we present the details of our proposed approach in the following subsections.}
\subsection{Databases}
In this study, \textcolor{black}{we selected five publicly available datasets}. The datasets include cases of WMHs resulting from both vascular and demyelinating origins. In addition, \textcolor{black}{we included one control dataset consisting of healthy aging brains to support model evaluation in non-pathological brain scans.} %one control dataset consisting of healthy aging brains was included to support model evaluation in non-pathological brain scans. 
\textcolor{black}{We used only FLAIR images due to best showing the WMH lesions.} \textcolor{black}{While some datasets included other MRI modalities such as T1- and T2-weighted sequences, WMHs appear hypointense on T1 and hyperintense on T2, making them less sensitive to subtle white abnormalities. FLAIR, by comparison, improves lesion contrast making it the most sensitive sequence for WMH \cite{wardlaw2015wmh}. } %Only the FLAIR scanners were used due to best showing the WMH lesions. 
Moreover, the datasets were acquired using 1.5T or 3T MRI scanners, varying in spatial resolution and scan dimensions. Table \ref{tab:datasets_summary} summarizes these characteristics for each dataset.
\begin{itemize}
\item \textbf{Datasets Containing Vascular-Origin WMHs:}
\textcolor{black}{We used a dataset from the \textbf{WMH Segmentation Challenge} \cite{WMHSegmentationChallenge} with lesions of presumed vascular origin.}
%A dataset from the \textbf{WMH Segmentation Challenge} \cite{WMHSegmentationChallenge} was used, whose lesions are presumed to be of vascular origin. 
\textcolor{black}{It incorporates 170 MRI scans from three institutes that used different scanners, providing a diverse set of acquisition conditions.}
%It consists of 170 MRI scans from three different institutes that used scanners from different vendors, which provided a diverse set of acquisition conditions.
 In addition, all the images underwent bias correction using SPM12 \cite{Kuijf_2019}. 

\textcolor{black}{We obtained another dataset from the \textbf{Utrecht Vascular Cognitive Impairment Study Group} \cite{diabetes}.}
%Another dataset with lesions of presumed vascular origin was obtained from the \textbf{Utrecht Vascular Cognitive Impairment Study Group} \cite{diabetes}. 
This dataset includes participants aged between 65 and 80 years, all of whom had type 2 diabetes for at least a year. The data comprises MRI scans from 60 patients and 54 control subjects without diabetes, resulting in \textcolor{black}{114} MRI scans. 

\item \textbf{Datasets Containing Demyelinating-Origin WMHs:}
Another type of WMH \textcolor{black}{lesion} of interest in this study is from demyelinating \textcolor{black}{origin}. The first dataset is \textbf{Brain MRI Dataset of MS with Consensus Manual Lesion Segmentation and Patient Meta Information} \cite{MS2}, published on Mendeley Data. It \textcolor{black}{comprises} scans from 60 patients aged between 15 to 56 years, covering  50 cases of \textcolor{black}{Relapsing-Remitting} MS (RRMS) and 10 from with Secondary Progressive MS (SPMS). 

The second dataset, from the \textbf{MSLesSeg ICPR 2024 competition} \cite{Rondinella2024, Rondinella2023Boosting, Rondinella2023Diffusion}, contains 53 scans from patients aged 18 and 59 years. \textcolor{black}{The dataset includes series acquired at multiple time points using various MRI scanners. }%The dataset includes series from various MRI scanners, acquired at at multiple time points using different MRI scanners. 
This dataset was preprocessed with anonymization, rigid registration to a 1mm³ MNI512 template and brain extraction.

\item \textbf{Control Dataset:}
\textcolor{black}{This study makes use of the \textbf{Neurocognitive aging data release with behavioral, structural and multi-echo functional MRI measures} dataset \cite{OpenNeuro}, published on OpenNeuro, as the control group.}
%The \textbf{Neurocognitive aging data release with behavioral, structural, and multi-echo functional MRI measures} dataset \cite{OpenNeuro}, published on OpenNeuro was used as the control group in this study.
This dataset \textcolor{black}{provides} MRI scans from 301 healthy adults, of which only a subset of \textcolor{black}{239 scans} with FLAIR images was selected. The participants were from a wide age range.
\end{itemize}
\begin{table*}[ht]
\centering
\small
\caption{Summary of datasets used in the study.}
\label{tab:datasets_summary}
\resizebox{\textwidth}{!}{%
\begin{tabular}{|l|c|c|c|c|}
\hline
\textbf{Dataset Name} & \textbf{Number of \textcolor{black}{scanners}} & \textbf{Type of scan} & \textbf{Volumes Size} & \textbf{Description / Origin} \\
\hline
WMH Segmentation Challenge \citep{WMHSegmentationChallenge} & 170 & 1.5T and 3T & 
\begin{tabular}[c]{@{}l@{}}
$240 \times 240 \times 48$ \\
$132 \times 256 \times 83$ \\
$128 \times 256 \times 103$ \\
$321 \times 240 \times 83$ \\
$256 \times 232 \times 48$ \\
\end{tabular} & Vascular-origin WMHs \\ \hline
Utrecht Vascular Cognitive Impairment Study \citep{diabetes} & \textcolor{black}{114} & 3T & 240 x 240 x 48 & Vascular-origin WMHs \\ \hline
Brain MRI Dataset of MS \citep{MS2} & 60 & Not mentioned & 256 x 256 x 23 & Demyelinating-origin WMHs \\\hline
MSLesSeg ICPR 2024 competition \cite{Rondinella2024, Rondinella2023Boosting, Rondinella2023Diffusion} & \textcolor{black}{93} & 1.5T & 182 x 218 x 182 & Demyelinating-origin WMHs \\\hline
Neurocognitive Aging Dataset \citep{OpenNeuro} & \textcolor{black}{239} & 3T & 256 × 256 × 46 & Healthy controls (no lesions) \\
\hline
\end{tabular}
}
\end{table*}
\subsection{Preprocessing}
\textcolor{black}{We developed a standardized preprocessing pipeline. It processes brain MRI slices from the defined datasets for lesion segmentation and classification}. The first step involves extracting the image slices of the 3D MRI volumes. Then, \textcolor{black}{we converted non-binary masks with irrelevant label values to binary by keeping only lesion-specific annotations}, discarding unrelated labels. After that,  Otsu’s method was used to generate a threshold mask isolating brain tissue from the background. This mask was refined via hole filling and was used to remove non-brain areas. Finally, all slices and masks were resized to a shape of 256×256 pixels using center cropping or symmetric zero-padding, depending on the original dimensions. For the control dataset, \textcolor{black}{we assigned binary lesion masks with only zeros} to each subject, indicating the absence of white matter lesions. Additionally, \textcolor{black}{we labeled slices containing lesions as 0 for vascular lesions and 1 for demyelinating lesions.}

\subsection{Segmentation Models}
\textcolor{black}{We used two} models for lesion segmentation: U-Net \citep{unet} and Attention U-Net. The Attention U-Net incorporates \textcolor{black}{Attention Gates} (AGs), which identify and emphasize relevant spatial regions in feature maps while suppressing irrelevant or noisy activations.

To further boost performance, \textcolor{black}{we integrated} additional attention modules into both U-Net and Attention U-Net: 
BAM and CBAM. BAM adds channel and spatial attention at the bottleneck of the network. \textcolor{black}{It} learns the regions and features to focus on or suppress, refining the intermediate \textcolor{black}{feature} maps accordingly. CBAM applies channel and spatial attention sequentially to adaptively refine features. These attention-enhanced modules have shown promising results in other domains like object detection and classification \citep{BAM, CBAM}. \textcolor{black}{They have also demonstrated effectiveness in medical image segmentation, including COVID-19 lesion delineation \citep{cbamapp} and organ segmentation tasks\textcolor{black}{,} such as cardiac structures \citep{bamapp}. We expect that their integration will improve WMH segmentation performance by enabling the network to better capture subtle intensity variations and spatial patterns, enhancing both lesion localization and boundary delineation.} 

While BAM and CBAM apply attention mechanisms either in parallel or sequentially across channel and spatial dimensions, attention gates compute attention coefficients for each pixel or spatial location using a grid attention approach. This gating signal is spatially aware and conditioned on local image context, allowing more precise focus on relevant regions.

Another reason why we selected these models was because the limited amount of available data made larger architectures, such as SAM or transformer-based models, less suitable.

\subsection{Feature Extraction}
\textcolor{black}{We first extracted}
individual lesions from segmentation masks and ground truth by identifying each connected region within the masks. For each individual lesion, \textcolor{black}{we extracted} a set of morphological and region-based features to quantitatively describe its shape, structure and spatial characteristics. These features include geometric descriptors such as area, perimeter, solidity, roundness, eccentricity, Feret angle, compactness, sphericity, aspect ratio and rectangularity. \textcolor{black}{They provide} insights into lesion size, elongation, orientation and compactness \cite{PETROVIC2020104027}. Additionally, \textcolor{black}{we computed} the relative position of each lesion’s centroid within the image. This extraction process was performed using the \texttt{uib\_vfeatures} library\footnote{Available at: https://github.com/explainingAI/uib\_vfeatures}.

\textcolor{black}{Radiologists interpret white matter lesions based on shape, size and location. For instance, geometric features such as eccentricity and aspect ratio capture lesion elongation, which aligns with the identification of ovoid periventricular lesions seen in multiple sclerosis. Roundness and compactness capture spherical lesions, which are attributed to microvascular disease \cite{radiologicalcriteria}. By linking these features to established radiological criteria, our analysis supports not only quantitative assessment but also improved clinical interpretability of the results.}

\subsection{Classification Models}

To classify each extracted lesion individually, we employed three machine learning models: Random Forest, Support Vector Machine (SVM) and Logistic Regression. 
\textcolor{black}{We decided to use shallow machine learning models because they can provide insight into how each feature contributes to predictions, which is critical in healthcare.}

% explanation of why

\section{Experiments}
\subsection{Experimental Setup}
The experiments were performed on a NVIDIA GeForce RTX 4090 GPU. They were implemented in Python language using version 3.10.13\textcolor{black}{,} PyTorch Framework 2.6.0 (CUDA 11.8) and core libraries such as Numpy, OpenCV, scikit-image and scikit-learn.
\subsection{Segmentation Approaches}
The experiments to train the segmentation models followed a series of approaches: 
\begin{itemize}
    \item 2D slices: The models were trained using 2D axial slices extracted from the 3D volumes. This approach treats each slice independently. 
    \item Patch-Based Training: \textcolor{black} {We divided each} 2D slice into smaller patches of size 64 × 64 pixels with a \textcolor{black}{16-pixel overlap, which were then fed into the network}. In an experiment, \textcolor{black}{we concatenated} the spatial position of each patch (row and column coordinates) to the bottleneck layer of the model. \textcolor{black}{This provided} additional positional information and \textcolor{black}{allowed us to} assess its impact on segmentation performance.
    \item 2.5D \textcolor{black}{Images}: This approach consists in the creation of three channel images by concatenating three adjacent 2D slices to provide context for the segmentation. The model outputs the segmentation mask of the middle slice. This approach helps to capture local 3D structural information while maintaining a lower computational complexity compared to full 3D models \citep{25Dimages}.   If one or both adjacent slices were unavailable, such as at volume boundaries, \textcolor{black}{we used} zero-filled tensors in their place. %https://arxiv.org/pdf/1812.08367
\end{itemize}

For the experiments using 2D slices and patch-based training, \textcolor{black}{we incorporated} the attention modules described in Section~\ref{sec:methods} into both U-Net and Attention U-Net. The experiments were conducted with and without the inclusion of BAM. Additionally,  \textcolor{black}{we integrated} CBAM into the encoder, the decoder, or both, depending on the specific experimental setup. 

\textcolor{black}{Models} using 2D slices were trained for 20 epochs. The 2.5D slice-based experiments were computationally more demanding and required 30 epochs to achieve optimal performance. In contrast, patch-based approaches required fewer training epochs, typically between 15 and 20. All experiments, regardless of the input strategy, used the Adam optimizer with a consistent learning rate of $1 \times 10^{-5}$ and \textcolor{black}{we used} Dice Loss as the loss function.

\subsection{Classification experiments}
\textcolor{black}{We extracted} the selected morphological and region-based features from both the predicted segmentation masks and the ground truth annotations. This setup allows us to evaluate the influence of segmentation accuracy on classification performance by comparing results obtained from predicted masks versus ground truth annotations. All features were standardized before training. \textcolor{black}{Models were initialized with a random state of 33 to ensure reproducibility.} The parameters used to train the machine learning classifiers \textcolor{black}{were obtained through a randomized search. They} are the following:

\begin{itemize}
    \item \textbf{Support Vector Machine (SVM)}: Implemented using a radial basis function (RBF) kernel with regularization parameter \( C = 10 \) and gamma \( = 0.1 \). To handle class imbalance, class weights were set to \texttt{balanced}. 
    \item \textbf{Logistic Regression (LR)}: Configured with L2 regularization with penalty parameter \( C = 1 \) and solved using the \texttt{lbfgs} optimizer. Class balancing was applied by setting the class weights to \texttt{balanced}. The model was trained for a maximum of 50 iterations.
    \item \textbf{Random Forest (RF)}: Consisted of an ensemble of 200 trees with a maximum depth of 20.
\end{itemize}

\subsection{Metrics}
\textcolor{black}{To evaluate each model, we employed different metrics depending on the task: 
\begin{itemize}
    \item For segmentation, we used  the Dice Coefficient and Jaccard Index. In addition, precision and recall were used to further assess performance. For models trained on individual 2D slices, \textcolor{black}{we reconstructed} the full 3D volume prior to metric computation. In models trained using patches, each 2D mask slice was reassembled from patches before reconstructing the 3D volume.
    \item For the classification task, we used precision and recall, along with accuracy and F1-score. Accuracy reflects the proportion of correctly classified instances, while the F1-score, as the harmonic mean of precision and recall, provides a balanced measure of performance.
\end{itemize}}

\subsection{Train and Validation split}
\textcolor{black}{We used five-fold cross-validation for all datasets. In each fold,} the data from each dataset \textcolor{black}{were split} into training \textcolor{black}{($\approx 80\%$)} and validation \textcolor{black}{($\approx 20\%$)} subsets. All slices from a given volume were kept within the same split to ensure consistency. The \textcolor{black}{approximate} number of scanners used \textcolor{black}{per fold} is detailed in Table~\ref{tab:train_val_split}. During training, batch sizes consisted of 25 images for both 2D and 2.5D slice-based models, whereas patch-based models used batches of 512 patches.
\begin{table}[ht]
\centering
\caption{\textcolor{black}{Approximate} training and validation scanners split \textcolor{black}{per fold} for each dataset.}
\label{tab:train_val_split}
\resizebox{1\linewidth}{!}{%
\begin{tabular}{|l|c|c|}
\hline
\textbf{Dataset Name} & \textbf{Training} & \textbf{Validation} \\
\hline
WMH Segmentation Challenge & 136 & 34 \\ \hline
Utrecht Vascular Cognitive Impairment Study & \textcolor{black}{91} & \textcolor{black}{23} \\\hline
Brain MRI Dataset of MS & 48 & 12 \\\hline
MSLesSeg ICPR 2024 competition & \textcolor{black}{74} & \textcolor{black}{19}\\\hline
Neurocognitive Aging Dataset & \textcolor{black}{191} & \textcolor{black}{48} \\
\hline
\end{tabular}
}
\end{table}

\section{Results and Discussion}
\subsection{Segmentation Results}
This section presents the results and discussion of the three segmentation approaches: 2D slices, patch-based training and 2.5D images.

\subsubsection{2D slices}
The results for 2D slice-based segmentation reveal several important trends. Table~\ref{tab:combined2dpatches} presents the \textcolor{black}{mean} performance metrics \textcolor{black}{across all folds for} this approach.  Among the models evaluated, the U-Net achieved \textcolor{black}{balanced} performance scores. The integration of BAM \textcolor{black}{did not lead to an improvement of} the segmentation results, \textcolor{black}{with} Dice \textcolor{black}{remaining comparable to the U-Net at $0.6727 \pm 0.037$}. \textcolor{black}{We observed a shift in Precision and Recall, with Precision decreasing to $0.6976 \pm 0.042$ and Recall increasing to $0.6825 \pm 0.037$}. This suggests that BAM \textcolor{black}{may} enhance the model's ability to focus on relevant features in the bottleneck layers \textcolor{black}{at the cost of introducing additional false positives}. 
\begin{table*}[ht]
\centering
\scriptsize
\caption{
\textcolor{black}{Mean and standard deviation of the} segmentation performance \textcolor{black}{on all datasets evaluated using five-fold cross validation for 2D slices and patches.} CBAM = Convolutional Block Attention Module, BAM = Bottleneck Attention Module.
}
\begin{tabular}{|l|c|c|c|c|c|c|c|c|}
\hline
\multirow{2}{*}{\textbf{Model}} & 
\multicolumn{4}{c|}{\textbf{2D Slices}} & 
\multicolumn{4}{c|}{\textbf{Patches}} \\
\cline{2-9}
& \textbf{Dice} & \textbf{Precision} & \textbf{Recall} & \textbf{Jaccard Index} 
& \textbf{Dice} & \textbf{Precision} & \textbf{Recall} & \textbf{Jaccard Index} \\
\hline
U-Net & $0.6736 \pm 0.027$ & $0.7280 \pm 0.034$ & $0.6604 \pm 0.030$ & $0.5772 \pm 0.029$ 
      & $0.5931 \pm 0.053$ & $0.6974 \pm 0.070$ & $0.5565 \pm 0.046$ & $0.4995 \pm 0.053$ \\\hline
U-Net + BAM & $0.6727 \pm 0.037$ & $0.6976 \pm 0.042$ & $0.6825 \pm 0.037$ & $0.5754 \pm 0.037$
            & $0.5943 \pm 0.006$ & $0.7079 \pm 0.017$ & $0.5510 \pm 0.008$ & $0.5001 \pm 0.007$ \\\hline
U-Net + BAM + CBAM (Encoder) &  $0.6638 \pm 0.063$ & $0.6950 \pm 0.055$ & $0.6769 \pm 0.068$ & $0.5676 \pm 0.063$ 
                             & $0.5848 \pm 0.047$ & $0.7046 \pm 0.062$ & $0.5400 \pm 0.041$ & $0.4907 \pm 0.046$ \\\hline
U-Net + BAM + CBAM (Decoder) & $0.6773 \pm 0.038$ & $0.7277 \pm 0.048$ & $0.6660 \pm 0.037$ & $0.5813 \pm 0.038$ 
                             & $0.6068 \pm 0.055$ & $0.7208 \pm 0.080$ & $0.5642 \pm 0.045$ & $0.5119 \pm 0.055$ \\\hline
U-Net + BAM + CBAM (Full) & $0.6881 \pm 0.017$ & $0.7525 \pm 0.031$ & $0.6692 \pm 0.017$ & $0.5918 \pm 0.019$
                          & $0.5770 \pm 0.016$ & $0.6989 \pm 0.026$ & $0.5314 \pm 0.021$ & $0.4821 \pm 0.018$ \\\hline
Attention U-Net & $0.6637 \pm 0.045$ & $0.7128 \pm 0.062$ & $0.6552 \pm 0.035$ & $0.5680 \pm 0.045$
                & $\mathbf{0.6134 \pm 0.018}$ & $\mathbf{0.7210 \pm 0.030}$ & $\mathbf{0.5741 \pm 0.012}$ & $\mathbf{0.5193 \pm 0.018}$\\\hline
Att. U-Net + BAM + CBAM (Full) & $\mathbf{0.7155 \pm 0.026}$ & $\mathbf{0.7681 \pm 0.030}$ & $\mathbf{0.7082 \pm 0.021}$ & $\mathbf{0.6199 \pm 0.026}$ 
                               & $0.5914 \pm 0.045$ & $0.6969 \pm 0.050$ & $0.5517 \pm 0.042$ & $0.4974 \pm 0.046$ \\
\hline
\end{tabular}
\label{tab:combined2dpatches}
\end{table*}
\begin{table*}[ht]
\centering
\small
\caption{
\textcolor{black}{Mean and standard deviation segmentation performance evaluated using five-fold cross-validation on} 2D slices and patches.
First row of each model is the mean of the combined 
Utrecht Vascular Cognitive Impairment Study and  WMH Segmentation Challenge, second row is the mean of MSLesSeg ICPR 2024 Competition and
Brain MRI Dataset of MS.
}
\resizebox{0.95\textwidth}{!}{%
\begin{tabular}{|l|c|c|c|c|c|c|c|c|}
\hline
\multirow{2}{*}{\textbf{Model}} &
\multicolumn{4}{c|}{\textbf{2D Slices}} &
\multicolumn{4}{c|}{\textbf{Patches}} \\
\cline{2-9}
& \textbf{Dice} & \textbf{Precision} & \textbf{Recall} & \textbf{Jaccard Index} 
  & \textbf{Dice} & \textbf{Precision} & \textbf{Recall} & \textbf{Jaccard Index} \\
\hline
U-Net 
 & $0.6992 \pm 0.014$ & $\mathbf{0.7378 \pm 0.031}$ & $0.6934 \pm 0.027$ & $0.5550 \pm 0.015$ 
 & $0.6877 \pm 0.013$ & $0.7761 \pm 0.025$ & $\mathbf{0.6447 \pm 0.022}$ & $0.5431 \pm 0.014$ \\
 & $0.5724 \pm 0.019$ & $0.6971 \pm 0.028$ & $0.5388 \pm 0.022$ & $0.4276 \pm 0.015$ 
 & $0.5180 \pm 0.021$ & $0.7422 \pm 0.030$ & $0.4509 \pm 0.027$ & $0.3817 \pm 0.017$ \\
\hline

U-Net + BAM 
 & $0.6966 \pm 0.018$ & $0.6845 \pm 0.029$ & $\mathbf{0.7411 \pm 0.014}$ & $0.5514 \pm 0.019$ 
 & $0.6816 \pm 0.013$ & $0.7865 \pm  0.028$ & $0.6291 \pm  0.025$ & $0.5363 \pm  0.013$ \\
 & $\mathbf{0.5849 \pm 0.021}$ & $0.6717 \pm 0.037$ & $\mathbf{0.5698 \pm 0.028}$ & $\mathbf{0.4381 \pm 0.018}$
 & $0.5158 \pm 0.026$ & $0.7515 \pm 0.030$ & $0.4384 \pm 0.030$ & $0.3786 \pm 0.022$ \\
\hline

U-Net + BAM + CBAM (Encoder) 
 & $0.6890 \pm 0.020$ & $0.6827 \pm 0.054$ & $0.7391 \pm 0.035$ & $0.5440 \pm 0.023$ 
 & $0.6673 \pm 0.009$ & $0.7888 \pm 0.042$ & $0.6080  \pm 0.023$ & $0.5206 \pm 0.009$ \\
 & $0.5627 \pm 0.010$ & $0.6628 \pm 0.067$ & $0.5521 \pm 0.047$ & $0.4192 \pm 0.011$
 & $0.5131 \pm 0.025$ & $0.7509 \pm 0.030$ & $0.4382  \pm 0.025$ & $0.3775 \pm 0.018$ \\
\hline

U-Net + BAM + CBAM (Decoder) 
 & $0.7014 \pm 0.016$ & $0.7353 \pm 0.052$ & $0.7047 \pm 0.037$ & $0.5578 \pm 0.018$
 & $0.6756 \pm 0.015$ & $0.7879 \pm 0.039$ & $0.6209 \pm  0.038$ & $0.5296 \pm 0.016$ \\
 & $0.5818 \pm 0.022$ & $0.6991 \pm 0.030$ & $0.5444 \pm 0.027$ & $0.4373 \pm 0.018$
 & $0.5249 \pm 0.021$ & $\mathbf{0.7544 \pm 0.052}$ & $0.4517 \pm 0.029$ & $0.3862 \pm 0.017$ \\
\hline

U-Net + BAM + CBAM (Full) 
 & $0.6830 \pm 0.015$ & $0.7357 \pm 0.038$ & $0.6697 \pm 0.033$ & $0.5369 \pm 0.016$
 & $0.6720 \pm 0.018$ & $\mathbf{0.7914 \pm 0.021}$ & $0.6090 \pm 0.026$ & $0.5243 \pm 0.019$ \\
 & $0.5569 \pm 0.016$ & $0.6972 \pm 0.041$ & $0.5135 \pm 0.022$ & $0.4141 \pm 0.013$
 & $0.5073 \pm 0.024$ & $0.7536 \pm 0.042$ & $0.4333 \pm 0.028$ & $0.3702 \pm 0.020$ \\
\hline

Attention U-Net 
 & $\mathbf{0.7095 \pm 0.015}$ & $0.7357 \pm 0.028$ & $0.7163 \pm 0.031$ & $\mathbf{0.5670 \pm 0.016}$
 & $0.6914 \pm 0.012$ & $0.7914 \pm 0.030$ & $0.6428 \pm 0.010$ & $\mathbf{0.5476 \pm 0.012}$ \\
 & $0.5783 \pm 0.007$ & $0.6992 \pm 0.047$ & $0.5458 \pm 0.023$ & $0.4336 \pm 0.008$
 & $\mathbf{0.5255 \pm 0.016}$ & $0.7483 \pm 0.023$ & $\mathbf{0.4561 \pm 0.018}$ & $0.3869\pm  0.012$ \\
\hline

Att. U-Net + Full BAM + CBAM 
 & $0.7028 \pm 0.013$ & $0.7197 \pm 0.029$ & $0.7222 \pm 0.011$ & $0.5597 \pm 0.015$
 & $\mathbf{0.6990 \pm 0.012}$ & $0.7839 \pm 0.032$ & $0.6416 \pm 0.019$ & $0.5444 \pm 0.013$ \\
 & $0.5732 \pm 0.019$ & $\mathbf{0.7141 \pm 0.024}$ & $0.5318 \pm 0.019$ & $0.4295 \pm 0.016$
 & $0.5250 \pm 0.015$ & $0.7469 \pm 0.025$ & $0.4536 \pm 0.015$ & $\mathbf{0.3879 \pm 0.013}$ \\
\hline

\end{tabular}
}
\label{tab:combined_slices_patches}
\end{table*}

The impact of CBAM varied depending on its placement within the network. Specifically, incorporating CBAM only in the encoder layers slightly decreased performance compared to the U-Net with BAM alone (Dice dropped from \textcolor{black}{$0.6727 \pm 0.037$ to $0.6638 \pm 0.063$}), indicating that attention in encoder layers alone might not optimally capture the necessary spatial and channel-wise features for segmentation. In contrast, adding CBAM to the decoder layers led to performance improvements (Dice = \textcolor{black}{$0.6773 \pm 0.038$}, Precision = \textcolor{black}{$0.7277\pm 0.048$}). Applying attention mechanisms throughout both the encoder and decoder, \textcolor{black}{achieved the best performance among U-Net variants}, with Dice = \textcolor{black}{$0.6881 \pm 0.017$}, Precision = \textcolor{black}{$0.7525\pm 0.031$}, Recall = \textcolor{black}{$0.6692 \pm 0.017$} and Jaccard Index = \textcolor{black}{$0.5918 \pm 0.019$, with low variability across folds}. 

\textcolor{black}{The Attention U-Net achieved mean performance comparable to the baseline U-Net, with relatively high standard deviations (Dice = $0.6637 \pm 0.045$, Precision = $0.7128 \pm 0.062$). This indicates more variability across folds}. However, extending the Attention U-Net with additional attention modules \textcolor{black}{achieved the highest overall mean} results \textcolor{black}{using 2D slices (Dice = $0.7155 \pm 0.026$, Precision = $0.7681 \pm 0.030$, Recall = $0.7082 \pm 0.021$, Jaccard Index = $0.6199 \pm 0.021$)}.

When examining the performance by lesion type, as shown in Table~\ref{tab:combined_slices_patches}, vascular lesions achieved better segmentation results compared to demyelinating lesions. The control dataset was not included in this analysis.

The \textcolor{black}{highest mean} results were achieved by the Attention U-Net, with a Dice score of \textcolor{black}{$0.7095 \pm 0.015$} and a Jaccard Index of \textcolor{black}{$0.5670 \pm 0.016$} in the Utrecht Vascular Cognitive Impairment Study and WMH Segmentation Challenge datasets. 

Performance decreased considerably on the MSLesSeg ICPR 2024 Competition and Brain MRI Dataset of MS datasets, with Dice scores dropping to around \textcolor{black}{0.55–0.58} for most models. The highest Dice score and Jaccard Index for this category were \textcolor{black}{$0.5849 \pm 0.021$} and \textcolor{black}{$0.4381 \pm 0.018$}, respectively, both achieved by the U-Net with BAM.  The U-Net with BAM and CBAM applied throughout the entire network (encoder and decoder) \textcolor{black}{achieved lower mean performance, despite achieving high overall metrics in the global analysis}. This contrast can be attributed to the inclusion of the control dataset in the global analysis, a dataset in which no lesions were present. This shows that, while full attention \textcolor{black}{models} performed well in scenarios with no pathology to detect, \textcolor{black}{they were} outperformed by other configurations when it came to identifying true pathological regions. \textcolor{black}{Although Attention U-Net achieved the highest mean scores for several metrics, the differences compared to other top-performing configurations fall within the reported standard deviations, indicating that the models exhibit comparable performance across folds.} 
\subsubsection{Patch-based training}
The results for 2D patch-based segmentation reveal that overall metrics were lower compared to 2D slice-based segmentation, as shown in Table \ref{tab:combined2dpatches}. The \textcolor{black}{highest mean} performance across all datasets was achieved by \textcolor{black}{Attention U-Net},  with Dice: \textcolor{black}{$0.6134 \pm 0.018$}, Precision: \textcolor{black}{$0.7210 \pm 0.030$}, Recall: \textcolor{black}{$0.5741 \pm 0.012$} and Jaccard Index: \textcolor{black}{$0.5193 \pm 0.018$}\textcolor{black}{; the differences relative to models containing BAM and CBAM}. 

As shown in Table~\ref{tab:combined_slices_patches}, vascular lesions again achieved better segmentation results compared to demyelinating lesions. The Attention U-Net \textcolor{black}{consistently} demonstrated strong performance in the Utrecht Vascular Cognitive Impairment Study and WMH Segmentation Challenge datasets, \textcolor{black}{being Attention U-Net with BAM and CBAM the highest mean Dice score}. Interestingly,  for the MSLesSeg ICPR 2024 Competition and Brain MRI Dataset of MS, \textcolor{black}{the Attention U-Net achieved a slightly higher Dice ($0.5255 \pm 0.016$}) compared to the U-Net with BAM.

To explore whether spatial context could improve patch-based segmentation, we conducted an \textcolor{black}{exploratory} experiment by concatenating patch position information at the Attention U-Net \textcolor{black}{bottleneck}, \textcolor{black}{yielding} the results from Table ~\ref{tab:segmentation_metrics}.
\begin{table}[!h]
\centering
\small
\caption{\textcolor{black}{Single-split} segmentation performance metrics across all datasets and lesion types adding position information at patch model.}
\begin{tabular}{|l|c|c|c|c|}
\hline
\textbf{Dataset} & \textbf{Dice} & \textbf{Precision} & \textbf{Recall} & \textbf{JI} \\
\hline
All Datasets    & 0.5933        & 0.6963             & 0.5531          & 0.5013       \\ \hline
Vascular Lesions               & \textbf{0.6931}        & \textbf{0.8250}             & \textbf{0.6233}          & \textbf{0.5501}       \\ \hline
Demyelinating Lesions          & 0.5244        & 0.7012             & 0.4733          & 0.3914       \\ 
\hline
\end{tabular}
\label{tab:segmentation_metrics}
\end{table}

While the inclusion of positional encoding slightly improved patch-based performance, particularly on \textcolor{black}{vascular} datasets, the results still fell short of those achieved by 2D slice-based models. Despite architectural improvements and the addition of spatial cues, patch-based segmentation remains less effective than 2D approaches for this task, especially in the context of complex or subtle lesion patterns. 

\subsubsection{2.5D input}
Following the analysis of 2D slice-based results, the two best overall models identified were: \textcolor{black} {Attention} U-Net with BAM and CBAM, \textcolor{black}{which achieved the highest global mean metrics}; and the Attention U-Net, which showed the \textcolor{black}{highest mean} performance on lesion-containing datasets. \textcolor{black}{We conducted an exploratory evaluation of} these two architectures using 2.5D input data to assess whether limited spatial context from adjacent slices improves segmentation performance. 

Table \ref{tab:dataset_metrics_comparison} shows each dataset performance. In the control dataset, the \textcolor{black}{Attention} U-Net + Full BAM + CBAM outperformed Attention U-Net, showing it can \textcolor{black}{accurately} identify healthy cases with no lesions. In vascular lesion datasets, \textcolor{black}{both models performed comparably, though Attention U-Net had better Dice, Precision and Jaccard Index on WMH,} indicating better sensitivity to lesion presence. For the demyelinating lesion datasets, both models struggled more, but Attention U-Net consistently scored higher in Dice, Precision, Recall and Jaccard Index, especially in ICPR. The global mean shows that \textcolor{black}{Attention} U-Net + Full BAM + CBAM \textcolor{black}{outperformed} Attention U-Net, mostly driven by the control dataset's \textcolor{black}{strong performance}.

Table \ref{tab:combined_dataset_metrics_comparison} shows the performance \textcolor{black}{by lesion type}. For the vascular lesion group, Attention U-Net \textcolor{black}{with BAM and CBAM} achieved \textcolor{black}{slightly} higher Dice \textcolor{black}{(0.6819 vs. 0.6743) and Recall (0.6325 vs. 0.6135)}, showing \textcolor{black}{marginally better vascular} lesion detection. In the demyelinating lesion group, Attention U-Net had better Dice and Recall, \textcolor{black}{suggesting} its better sensitivity \textcolor{black}{to demyelinating lesions}.

Compared to the 2D slice-based experiments, both models showed a decrease in overall performance with 2.5D input. However, they still performed better than patch-based models, although not by a significant margin. This suggests that while 2.5D inputs provide additional contextual information, they may also introduce noise or complexity that the current models do not fully leverage.

\begin{figure}[!t]
\centering
\includegraphics[scale=0.35]{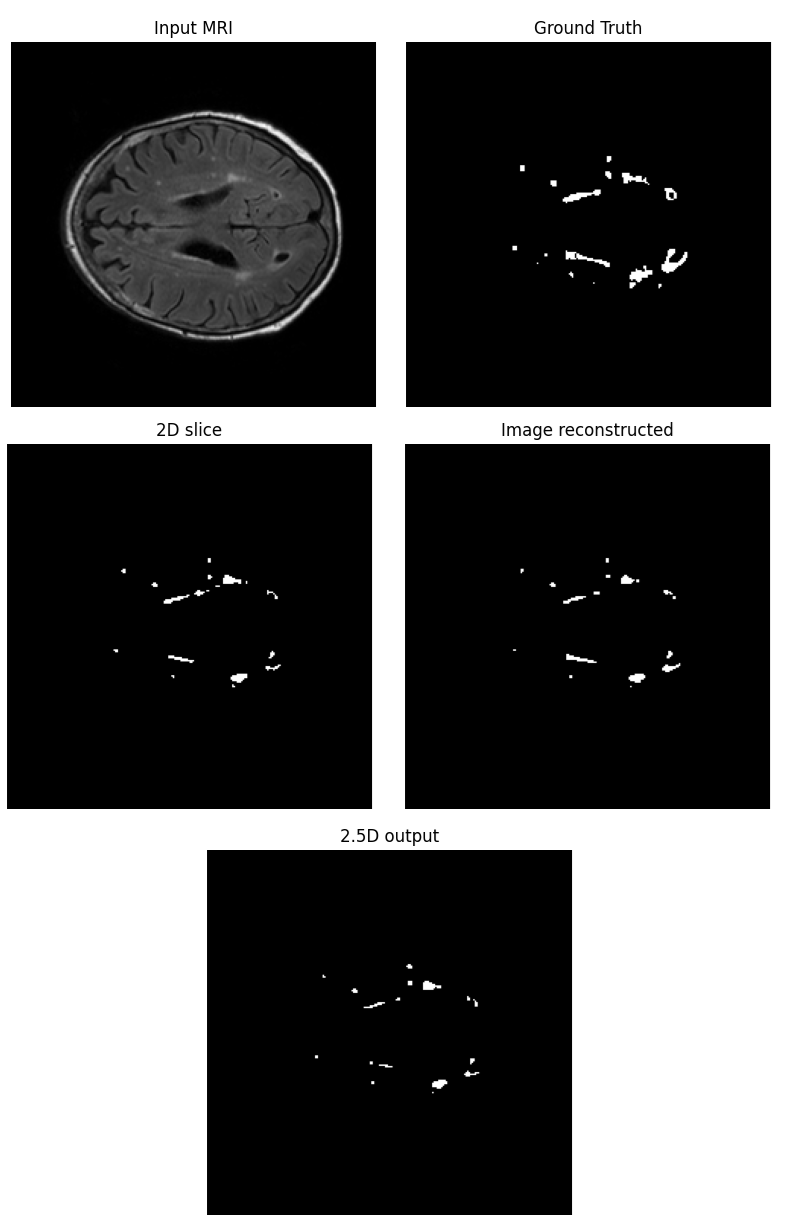}
\caption{This figure shows the results of segmentation using 2D slices, patches and 2.5D on WMHs of vascular origin with Attention U-Net.}
\label{fig:segmentation_results}
\end{figure}

In general, 2D slice-based models outperformed both 2.5D and patch-based models in this study. \textcolor{black}{Attention} U-Net + Full BAM + CBAM, \textcolor{black}{demonstrated strong performance} particularly on healthy/control cases and achieves solid performance across metrics. Meanwhile, the Attention U-Net, delivers greater lesion sensitivity, especially for vascular lesions. \textcolor{black}{Incorporating attention modules, particularly when applied throughout the network, improved U-Net performance,} while Attention U-Net proved \textcolor{black}{effective} for lesion segmentation. Figure~\ref{fig:segmentation_results} illustrates segmentation differences on WMHs using 2D slices, patches and 2.5D approaches with the Attention U-Net.

\textcolor{black}{Our findings are consistent with recent evidence that real-time AI segmentation can enhance diagnostic efficiency and accuracy \citep{realtimesegmentation}, highlighting the clinical value of our approach that emphasizes interpretable lesion localization and the extraction of radiologically meaningful features.}

Although the inclusion of five public datasets was intended to improve model generalizability, we acknowledge that dataset heterogeneity (such as differences in scanner models, acquisition parameters and demographic composition) can introduce domain shifts that affect performance. Similar concerns have been highlighted in recent work exploring dataset variability in AI-driven diagnostic systems and its implications for generalizability \citep{retinal}.

\begin{table*}[!ht]
\small
\centering
\caption{\textcolor{black}{Single-split segmentation performance metrics} per-dataset: 2.5D Attention U-Net vs 2.5D \textcolor{black}{Attention U-Net + Full BAM + CBAM}.}
\begin{tabular}{|l|c|c|c|c|c|c|c|c|}
\hline
\multirow{2}{*}{\textbf{Dataset}} & \multicolumn{4}{c|}{\textbf{Attention U-Net}} & \multicolumn{4}{c|}{\textbf{\textcolor{black}{Attention U-Net + Full BAM + CBAM}}} \\
\cline{2-9}
 & \textbf{Dice} & \textbf{Precision} & \textbf{Recall} & \textbf{Jaccard Index} & \textbf{Dice} & \textbf{Precision} & \textbf{Recall} & \textbf{Jaccard Index} \\
\hline
Control & 0.6042 & 0.6042 & 0.6042 & 0.6042 & 
\textbf{0.8333} & \textbf{0.8333} & \textbf{0.8333} & \textbf{0.8333} \\
\hline
Utrecht & 0.5919 & {0.7524} & 0.5115 & 0.4386 & 
\textbf{0.6118} & 0.7508 & \textbf{0.5441} & \textbf{0.4561} \\
\hline
WMH & \textbf{0.7569} & \textbf{0.8275} & 0.7155 & \textbf{0.6214} & 
0.7521 & 0.8155 & \textbf{0.7209} & 0.6158\\
\hline
ICPR & \textbf{0.6973} & \textbf{0.7605} & \textbf{0.6618} & \textbf{0.5507} & 0.6743 & 0.7579 & 0.6321 & 0.5271 \\
\hline
BrainMRI & 0.4032 & 0.5727 & \textbf{0.3782} & 0.2701 & 
\textbf{0.4082} & \textbf{0.5873} & 0.3774 & \textbf{0.2746} \\
\hline
\textbf{Global Mean} & 0.6107 & 0.7035 & 0.5742 & 0.4970 & 
\textbf{0.6596} & \textbf{0.7490} & \textbf{0.6226} & \textbf{0.5414} \\
\hline
\end{tabular}
\label{tab:dataset_metrics_comparison}
\end{table*}

\begin{table*}[ht]
\centering
\small
\caption{\textcolor{black}{Single-split} segmentation performance on combined datasets: 2.5D Attention U-net vs \textcolor{black}{Attention U-Net + Full BAM + CBAM }.}
\begin{tabular}{|l|c|c|c|c|c|c|c|c|}
\hline
\multirow{2}{*}{\textbf{Dataset Combination}} & \multicolumn{4}{c|}{\textbf{Attention U-Net}} & \multicolumn{4}{c|}{\textbf{\textcolor{black}{Attention U-Net + Full BAM + CBAM }}} \\
\cline{2-9}
 & \textbf{Dice} & \textbf{Precision} & \textbf{Recall} & \textbf{Jaccard Index} & \textbf{Dice} & \textbf{Precision} & \textbf{Recall} & \textbf{Jaccard Index} \\
\hline
Utrecht + WMH & 0.6743 & \textbf{0.7900} & 0.6135 & 0.5300 & 
\textbf{0.6819} & 0.7832 & \textbf{0.6325} & \textbf{0.5359} \\ \hline
ICPR + BrainMRI & \textbf{0.5502} & 0.6666 & \textbf{0.5200} & \textbf{0.4104} 
& 0.5413 & \textbf{0.6726} & 0.5048 & 0.4009 \\
\hline
\end{tabular}
\label{tab:combined_dataset_metrics_comparison}
\end{table*}

\textcolor{black}{\subsection{Classification Results}}
\textcolor{black}{We obtained the segmentation masks using the Attention U-Net model}, which achieved the best results for lesion segmentation. Only slices containing at least one lesion were included in the training.
\begin{table*}[!t]
\small
\centering
\caption{Classification results using ground truth vs. segmentation masks.}
\label{tab:gt_vs_seg}
\resizebox{\textwidth}{!}{%
\begin{tabular}{|l|ccc|ccc|c|ccc|ccc|c|}
\hline
\multirow{2}{*}{\textbf{Model}} 
& \multicolumn{6}{c|}{\textbf{Ground Truth}} 
& \multirow{2}{*}{\textbf{Acc.}} 
& \multicolumn{6}{c|}{\textbf{Segmentation Masks}} 
& \multirow{2}{*}{\textbf{Acc.}} \\
\cline{2-7} \cline{9-14}
& \multicolumn{3}{c|}{\textbf{Vascular}} & \multicolumn{3}{c|}{\textbf{Demyelinating}} 
& 
& \multicolumn{3}{c|}{\textbf{Vascular}} & \multicolumn{3}{c|}{\textbf{Demyelinating}} 
& \\
& Precision & Recall & F1 & Precision & Recall & F1 
& 
& Precision & Recall & F1 & Precision & Recall & F1 
& \\
\hline
SVM & \textbf{0.75} & 0.63 & 0.68 & 0.42 & 0.56 & 0.48 & 0.61 
    & \textbf{0.71} & 0.68 & 0.69 & \textbf{0.39} & 0.43 & 0.41 & 0.60 \\ \hline
LR  & 0.75 & 0.55 & 0.64 & 0.41 & \textbf{0.63} & \textbf {0.49} & 0.58 
    & 0.71 & 0.60 & 0.65 & 0.36 & \textbf{0.48} & \textbf{0.41} & 0.56 \\ \hline
RF  & 0.73 & \textbf{0.71} & \textbf{0.72} & \textbf{0.44} & 0.46 & 0.45 & \textbf{0.63} 
    & 0.70 & \textbf{0.71} & \textbf{0.71} & 0.37 & 0.37 & 0.37 & \textbf{0.60} \\
\hline
\end{tabular}
}
\end{table*}

Table \ref{tab:gt_vs_seg} shows the performance results of the classification. Across all models, a slight decrease in overall accuracy is observed when using segmentation masks instead of ground truth annotations. This drop is not substantial, demonstrating that even though the segmentation is not perfect, it achieves similar results to using ground truth masks.

SVM shows a more stable performance across vascular and demyelinating classes. Logistic Regression shows weaker performance compared to both SVM and Random Forest. Random Forest is the most robust model overall, with the highest F1-score for the Vascular class in both ground truth and segmentation mask conditions. The Vascular class shows high precision and recall across all models and input types, which aligns with the segmentation results, indicating that vascular features are well captured. The demyelinating class is harder to classify, showing lower precision and recall across all models and being more susceptible to performance drops when using segmentation masks.

In line with these observations, our approach moves beyond CNN-based feature extraction \citep{deepbelif}, by integrating attention-based segmentation with radiological feature extraction, emphasizing the advantages of combining deep learning with interpretable, feature-based analyses.

\section{Conclusions}\label{sec:conclusiones}
This study evaluated WMHs segmentation and classification in brain MRI for vascular and demyelinating pathologies. \textcolor{black}{We compared} 2D slices, patch-based and 2.5D inputs with attention-enhanced architectures across diverse datasets to assess their effectiveness in different clinical scenarios.

Our results demonstrate that 2D slice-based segmentation consistently outperformed both patch-based and 2.5D approaches across most datasets and metrics. The integration of attention modules \textcolor{black}{throughout the network slightly} improved \textcolor{black}{model performance}, with the \textcolor{black}{Attention} U-Net augmented by BAM and CBAM across both encoder and decoder layers achieving the highest overall segmentation performance, particularly in healthy subjects. In contrast, Attention U-Net showed \textcolor{black}{higher} lesion-specific sensitivity, achieving \textcolor{black}{improved} Dice and Recall scores in lesion-containing datasets. \textcolor{black}{In general, results suggest that while the Attention U-Net is a strong and consistent performer, multiple architectures achieve comparable performance, when variability across folds is taken into account.}

Despite these results, our approach has some limitations. The available dataset was relatively small and highly heterogeneous, incorporating scans acquired from different institutions using varying scanner models and imaging protocols.
In the classification task, morphological and region-based features extracted from segmentation masks proved effective for \textcolor{black}{discriminating}
lesion types. The three classifiers (SVM, Logistic Regression and Random Forest) achieved consistent results. Performance varied only slightly between ground truth and predicted segmentations. This suggests that minor segmentation errors did not significantly affect classification accuracy. 

Overall, this work highlights the role of attention mechanisms in lesion segmentation, the effectiveness of 2D slice-based methods and the value of shape-based features for lesion classification.

In future work, we aim to expand our research through the integration of multimodal data, following previous studies that have highlighted the value of combining imaging modalities and clinical information \cite{multimodal}. 
Additionally, we consider studying 3D shape descriptors \citep{3ddescriptors} by stacking the 2D segmented slices to form a 3D reconstruction, which may allow for a more detailed characterization of lesion morphology. 
Given the variability across datasets, we will focus on strategies to mitigate dataset heterogeneity, as recent studies have begun to offer solutions to this challenge to improve robustness \textcolor{black}{\citep{domainshift, query-guided}}.
\section*{Acknowledgments}
This work was supported by the following grants:  Project PID2019-104829RA-I00 ‘‘EXPLainable Artificial INtelligence systems for health and well-beING (EXPLAINING)’’ funded by MCIN/AEI, Spain/10.13039/501100011033; Project PID2023-149079OB-I00 funded by MICIU/AEI, Spain/10.13039/5011000
11033/ and ERDF, EU and Project PID2022-136779OB-C32 (PLEISAR) funded by MICIU/ AEI /10.13039/501100011033/ and FEDER, EU

\bibliographystyle{elsarticle-num}
\bibliography{refs}

@article{wardlaw2015wmh,
 author = {Wardlaw, J. M. and Valdés Hernández, M. C. and Muñoz-Maniega, S.},
  title = {What are white matter hyperintensities made of? Relevance to vascular cognitive impairment},
  journal = {Journal of the American Heart Association},
  year = {2015},
  month = jun,
  day = {23},
  volume = {4},
  number = {6},
  pages = {e001140},
  doi = {10.1161/JAHA.114.001140},
  note = {Erratum in: J Am Heart Assoc. 2016 Jan 13;5(1):e002006. doi: 10.1161/JAHA.115.002006},
  pmid = {26104658},
  pmcid = {PMC4599520}
}

@article{corneal,
  author    = {G. Ali and Marwa M. Eid and Omar G. Ahmed and Mostafa Abotaleb and Anas M. Zein Alaabdin and Bosco Apparatus Buruga},
title={Artificial intelligence in Corneal Topography: A Short Article in Enhancing Eye Care}, volume={2023}, 
  url     = {https://doi.org/10.58496/MJAIH/2023/006},

DOI={10.58496/MJAIH/2023/006},  
journal={Mesopotamian Journal of Artificial Intelligence in Healthcare}, year={2023}, 
month={Jun.}, 
pages={31–34} 
}

@article{bamapp,
  title     = {DSBAV-Net: Depthwise Separable Bottleneck Attention V-Shaped Network with Hybrid Convolution for Left Atrium Segmentation},
  author    = {Ocal, H.},
  journal   = {Arabian Journal for Science and Engineering},
  volume    = {50},
  pages     = {1097--1108},
  year      = {2025},
  doi       = {10.1007/s13369-024-09131-1},
  url       = {https://doi.org/10.1007/s13369-024-09131-1}
}

@article{cbamapp,
  title     = {MIDCAN: A multiple input deep convolutional attention network for Covid-19 diagnosis based on chest CT and chest X-ray},
  author    = {Yu-Dong Zhang and Zheng Zhang and Xin Zhang and Shui-Hua Wang},
  journal   = {Pattern Recognition Letters},
  volume    = {150},
  pages     = {8--16},
  year      = {2021},
  issn      = {0167-8655},
  doi       = {10.1016/j.patrec.2021.06.021},
url = {https://www.sciencedirect.com/science/article/pii/S0167865521002270},

}

@article{zhang2023white,
  author = {Zhang, L. J. and Tian, D. C. and Yang, L. and Shi, K. and Liu, Y. and Wang, Y. and Shi, F. D.},
  title = {White matter disease derived from vascular and demyelinating origins},
  journal = {Stroke and Vascular Neurology},
  year = {2024},
  month = aug,
  day = {27},
  volume = {9},
  number = {4},
  pages = {344--350},
  doi = {10.1136/svn-2023-002791},
  pmid = {37699727},
  pmcid = {PMC11420911}
}

@article{datasetsdesc,
   author  = {Al-Quraishi, T. and Ng, C.K. and Mahdi, O.A. and Gyasi, A. and Al-Quraishi, N.},
  title   = {Advanced ensemble classifier techniques for predicting tumor viability in osteosarcoma histological slide images},
  journal = {Applied Data Science and Analysis},
  year    = {2024},
  pages   = {52--68},
  doi     = {https://doi.org/10.58496/ADSA/2024/006}

}

@dataset{OpenNeuro,
  author       = {R.N. Spreng and R. Setton and U. Alter and B.N. Cassidy and B. Darboh and E. DuPre and K. Kantarovich and A.W. Lockrow and L. Mwilambwe-Tshilobo and W.-M. Luh and P. Kundu and G.R. Turner},
  title        = {Neurocognitive aging data release with behavioral, structural, and multi-echo functional MRI measures},
  year         = {2022},
  howpublished = {[dataset] OpenNeuro, v1.0.13},
  doi = {doi:10.18112/openneuro.ds003592.v1.0.13}
}

@misc{WMHSegmentationChallenge,
  author    = {H. Kuijf and M. Biesbroek and J. de Bresser and R. Heinen and C. Chen and W. van der Flier and F. Barkhof and M. Viergever and G.J. Biessels},
  title     = {Data of the White Matter Hyperintensity (WMH) Segmentation Challenge [dataset]},
  publisher = {DataverseNL},
  year      = {2022},
  note      = {{DataverseNL},v1},
  doi       = {10.34894/AECRSD},
  url       = {https://doi.org/10.34894/AECRSD}
}

@misc{diabetes,
author    = {H. Kuijf and G. J. Biessels},
  title     = {Data of White matter hyperintensity shape and location feature analysis on brain MRI; proof of principle study in patients with diabetes [dataset]},
  year      = {2023},
  note      = {{DataverseNL},v1},
  doi       = {10.34894/KG5WBO},
  url       = {https://doi.org/10.34894/KG5WBO}
}

@misc{MS2,
 author    = {M. Muslim and A. Ali},
  title     = {Brain MRI Dataset of Multiple Sclerosis with Consensus Manual Lesion Segmentation and Patient Meta Information [dataset]},
  year      = {2022},
  note      = {{Mendeley Data}, v1},
  doi       = {10.17632/8bctsm8jz7.1},
  url       = {https://doi.org/10.17632/8bctsm8jz7.1}
}

@inbook{Rondinella2024,
  author    = {A. Rondinella and F. Guarnera and E. Crispino and G. Russo and C. Di Lorenzo and D. Maimone and F. Pappalardo and S. Battiato},
  title     = {ICPR 2024 Competition on Multiple Sclerosis Lesion Segmentation—Methods and Results},
  booktitle = {2024 International Conference on Pattern Recognition (ICPR)},
  year      = {2024},
  publisher = {IEEE},
  year = {2024},
  month = nov,
  pages = {1–16}
}

@article{Rondinella2023Boosting,
  author    = {A. Rondinella and E. Crispino and F. Guarnera and O. Giudice and A. Ortis and G. Russo and C. Di Lorenzo and D. Maimone and F. Pappalardo and S. Battiato},
  title     = {Boosting multiple sclerosis lesion segmentation through attention mechanism},
  journal   = {Computers in Biology and Medicine},
  volume    = {161},
  year      = {2023},
  pages     = {107021},
  doi       = {10.1016/j.compbiomed.2023.107021},
  url       = {https://doi.org/10.1016/j.compbiomed.2023.107021}
}

@inproceedings{Rondinella2023Diffusion,
  author    = {A. Rondinella and F. Guarnera and O. Giudice and A. Ortis and G. Russo and E. Crispino and F. Pappalardo and S. Battiato},
  title     = {Enhancing Multiple Sclerosis Lesion Segmentation in Multimodal MRI Scans with Diffusion Models},
  booktitle = {2023 IEEE International Conference on Bioinformatics and Biomedicine Workshops (CMISF)},
  year      = {2023},
  publisher = {IEEE}
}

@article{Kuijf_2019,
author    = {H. J. Kuijf and M. W. M. Biesbroek and J. de Bresser and R. Heinen and C. L. H. Chen and W. M. van der Flier and F. Barkhof and M. A. Viergever and G. J. Biessels},
  title     = {Standardized Assessment of Automatic Segmentation of White Matter Hyperintensities and Results of the WMH Segmentation Challenge},
  journal   = {IEEE Transactions on Medical Imaging},
  volume    = {38},
  number    = {11},
  pages     = {2556--2568},
  year      = {2019},
  month     = {Nov},
  doi       = {10.1109/TMI.2019.2905770},
  publisher = {IEEE}
}

@inproceedings{unet,
  author    = {Olaf Ronneberger and Philipp Fischer and Thomas Brox},
  title     = {U-Net: Convolutional Networks for Biomedical Image Segmentation},
  booktitle = {Medical Image Computing and Computer-Assisted Intervention -- MICCAI 2015},
  series    = {Lecture Notes in Computer Science},
  volume    = {9351},
  pages     = {234--241},
  year      = {2015},
  publisher = {Springer},
  url       = {https://arxiv.org/abs/1505.04597},
  note      = {Available at arXiv:1505.04597 [cs.CV]}
}

@misc{attentionunet,
      title={Attention U-Net: Learning Where to Look for the Pancreas}, 
      author={Ozan Oktay and Jo Schlemper and Loic Le Folgoc and Matthew Lee and Mattias Heinrich and Kazunari Misawa and Kensaku Mori and Steven McDonagh and Nils Y Hammerla and Bernhard Kainz and Ben Glocker and Daniel Rueckert},
      year={2018},
      eprint={1804.03999},
      archivePrefix={arXiv},
      primaryClass={cs.CV},
      url={https://arxiv.org/abs/1804.03999}, 
}

@misc{BAM,
      title={BAM: Bottleneck Attention Module}, 
      author={Jongchan Park and Sanghyun Woo and Joon-Young Lee and In So Kweon},
      year={2018},
      eprint={1807.06514},
      archivePrefix={arXiv},
      primaryClass={cs.CV},
      url={https://arxiv.org/abs/1807.06514}, 
}

@misc{CBAM,
      title={CBAM: Convolutional Block Attention Module}, 
      author={Sanghyun Woo and Jongchan Park and Joon-Young Lee and In So Kweon},
      year={2018},
      eprint={1807.06521},
      archivePrefix={arXiv},
      primaryClass={cs.CV},
      url={https://arxiv.org/abs/1807.06521}, 
}

@misc{25Dimages,
  title        = {2.5D Deep Learning for CT Image Reconstruction using a Multi-GPU implementation},
  author       = {Amirkoushyar Ziabari and Dong Hye Ye and Somesh Srivastava and Ken D. Sauer and Jean-Baptiste Thibault and Charles A. Bouman},
  year         = {2018},
  eprint       = {1812.08367},
  archivePrefix= {arXiv},
  primaryClass = {eess.IV},
  url          = {https://arxiv.org/abs/1812.08367},
  note         = {arXiv preprint arXiv:1812.08367}
}

@article{PETROVIC2020104027,
 title     = {Sickle-cell disease diagnosis support selecting the most appropriate machine learning method: Towards a general and interpretable approach for cell morphology analysis from microscopy images},
  author    = {Nataša Petrović and Gabriel Moyà-Alcover and Antoni Jaume-i-Capó and Manuel González-Hidalgo},
  journal   = {Computers in Biology and Medicine},
  volume    = {126},
  pages     = {104027},
  year      = {2020},
  issn      = {0010-4825},
  doi       = {10.1016/j.compbiomed.2020.104027},
  url       = {https://www.sciencedirect.com/science/article/pii/S0010482520303589}
}

@article{radiologicalcriteria,
  author    = {S. Medrano Martorell and M. Cuadrado Blázquez and D. García Figueredo and S. González Ortiz and J. Capellades Font},
  title     = {Hyperintense punctiform images in the white matter: A diagnostic approach},
  journal   = {Radiología (English Edition)},
  volume    = {54},
  number    = {4},
  pages     = {321--335},
  year      = {2012},
  issn      = {2173-5107},
  doi       = {10.1016/j.rxeng.2011.09.001},
  url       = {https://doi.org/10.1016/j.rxeng.2011.09.001}
}

@article{realtimesegmentation,
  author    = {Shin, J.},
  title     = {Revolutionizing Medical Imaging with Artificial Intelligence: Real-Time Segmentation for Enhanced Diagnostics},
  journal   = {EDRAAK},
  year      = {2024},
  pages     = {18--25},
  doi       = {10.70470/EDRAAK/2024/003}
}

@article{retinal,
  author    = {Yang, Y. and Wang, H. and Ji, C. and Niu, Y.},
  title     = {Artificial Intelligence-Driven Diagnostic Systems for Early Detection of Diabetic Retinopathy: Integrating Retinal Imaging and Clinical Data},
  journal   = {SHIFAA},
  volume    = {2023},
  pages     = {83--90},
  year      = {2023},
  month     = {Jul},
  doi       = {10.70470/SHIFAA/2023/010}
}

@article{query-guided,
title = {Query-guided generalizable medical image segmentation},
journal = {Pattern Recognition Letters},
volume = {184},
pages = {52-58},
year = {2024},
issn = {0167-8655},
doi = {https://doi.org/10.1016/j.patrec.2024.06.005},
author = {Zhiyi Yang and Zhou Zhao and Yuliang Gu and Yongchao Xu}
}

@article{deepbelif,
  author    = {Sheela, M. and Amirthayogam, G. and Hephzipah, J. J. and Suganthi, R. and Karthikeyan, T. and Gopianand, M.},
  title     = {Advanced Brain Tumor Classification Using DEEPBELEIF-CNN Method},
  journal   = {Babylonian Journal of Machine Learning},
  volume    = {2024},
  pages     = {89--101},
  year      = {2024},
  doi       = {10.58496/BJML/2024/009},
  url       = {https://doi.org/10.58496/BJML/2024/009}
}

@article{multimodal,
  title        = {The Benefit of Artificial Intelligence in the Analysis of Malignant Brain Diseases: A Mini Review},
  author       = {Al-Shahwani, H. I. W. and Faieq, A. K.},
  year         = {2023},
  journal      = {Mesopotamian Journal of Artificial Intelligence in Healthcare},
  pages        = {57--60},
  doi          = {10.58496/MJAIH/2023/011}
}

@article{3ddescriptors,
  author={M. A. Westenberg and J. B. T. M. Roerdink and M. H. F. Wilkinson},
  journal={IEEE Transactions on Image Processing}, 
  title={Volumetric Attribute Filtering and Interactive Visualization Using the Max-Tree Representation}, 
  year={2007},
  volume={16},
  number={12},
  pages={2943-2952},
  doi={10.1109/TIP.2007.909317}
}

@article{domainshift,
  author    = {Chen, Xiaocong and Li, Yun and Yao, Lina and Adeli, Ehsan and Zhang, Yu and Wang, Xianzhi},
  title     = {Generative adversarial U-Net for domain-free few-shot medical diagnosis},
  journal   = {Pattern Recognition Letters},
  volume    = {157},
  pages     = {112--118},
  year      = {2022},
  issn      = {0167-8655},
  doi       = {10.1016/j.patrec.2022.03.022},
  url       = {https://www.sciencedirect.com/science/article/pii/S0167865522000873}
}

\end{document}